\documentclass[11pt]{article}
\usepackage[final]{eacl2023}
\usepackage{amsmath}
\usepackage{bm}
\usepackage{times}
\usepackage{latexsym}
\usepackage[T1]{fontenc}
\usepackage[utf8]{inputenc}
\usepackage{multirow,tabularx}
\usepackage{microtype}
\usepackage{inconsolata}
\usepackage{enumitem}
\usepackage{todonotes}
\usepackage{tcolorbox}
\usepackage{color,soul}
\usepackage{hyperref}
\usepackage{flushend}
\usepackage[section]{placeins}
\usepackage{epstopdf}
\usepackage{multirow,tabularx}
\usepackage{pbox}
\usepackage{cleveref}
\usepackage{algorithm}
\usepackage{algpseudocode}
\usepackage{svg}

\crefformat{section}{\S#2#1#3}
\crefformat{subsection}{\S#2#1#3}
\crefformat{subsubsection}{\S#2#1#3}
\crefrangeformat{section}{\S\S#3#1#4 to~#5#2#6}
\crefmultiformat{section}{\S\S#2#1#3}{ and~#2#1#3}{, #2#1#3}{ and~#2#1#3}

\RequirePackage{colortbl}
\definecolor{airforceblue}{rgb}{0.36, 0.54, 0.66}
\definecolor{amaranth}{rgb}{0.9, 0.17, 0.31}
\definecolor{applegreen}{rgb}{0.55, 0.71, 0.0}
\definecolor{alizarin}{rgb}{0.82, 0.1, 0.26}
\definecolor{azure}{rgb}{0.0, 0.5, 1.0}
\definecolor{cadmiumgreen}{rgb}{0.0, 0.42, 0.24}

\title{\textsc{MUTANT}: A Multi-sentential Code-mixed Hinglish Dataset}

\author{Rahul Gupta \\
  IIT Gandhinagar\\ Gandhinagar, Gujarat, India \\
  \texttt{gupta.rahul@iitgn.ac.in} \\\And
  Vivek Srivastava \\
  TCS Research\\ Pune, Maharashtra, India \\
  \texttt{srivastava.vivek2@tcs.com} \\\And
  Mayank Singh \\
  IIT Gandhinagar\\ Gandhinagar, Gujarat, India \\
  \texttt{singh.mayank@iitgn.ac.in} \\}

\begin{document}
\maketitle

\begin{abstract}
The multi-sentential long sequence textual data unfolds several interesting research directions pertaining to natural language processing and generation. Though we observe several high-quality long-sequence datasets for English and other monolingual languages, there is no significant effort in building such resources for code-mixed languages such as Hinglish (code-mixing of Hindi-English). In this paper, we propose a novel task of identifying multi-sentential code-mixed text (MCT) from multilingual articles. As a use case, we leverage multilingual articles from two different data sources and build a first-of-its-kind multi-sentential code-mixed Hinglish dataset i.e., \textsc{MUTANT}. We propose a token-level language-aware pipeline and extend the existing metrics measuring the degree of code-mixing to a multi-sentential framework and automatically identify MCT in the multilingual articles. The \textsc{MUTANT} dataset comprises 67k articles with 85k identified Hinglish MCTs. To facilitate future research, we make the \href{https://drive.google.com/file/d/1CLs2E6C3ygbi3eJ5IvH4GbN4EE0mli5O/view?usp=sharing}{dataset} publicly available. 
\end{abstract}

\section{Introduction}
\label{sec: intro}
Over the years, we have seen enormous downstream applications of multi-sentential datasets in the areas such as question-answering \cite{joshi2017triviaqa, tapaswi2016movieqa}, summarization \cite{sharma2019bigpatent, cachola2020tldr}, machine translation \cite{bao2021g}, etc. The existing state-of-the-art methods prove challenging to scale effectively and efficiently on multi-sentential long sequence text \cite{ainslie2020etc}, which unplugs several exciting research avenues. Unfortunately, to a large extent, the majority of the research on multi-sentential data is dominated by a few popular monolingual languages such as English, Chinese, and Spanish. Due to this, code-mixed languages (among other low-resource and under-explored languages) suffer from non-existent works in the aforementioned areas of interest. 
\begin{figure}[t]
\centering
\includegraphics[width=1.0\linewidth]{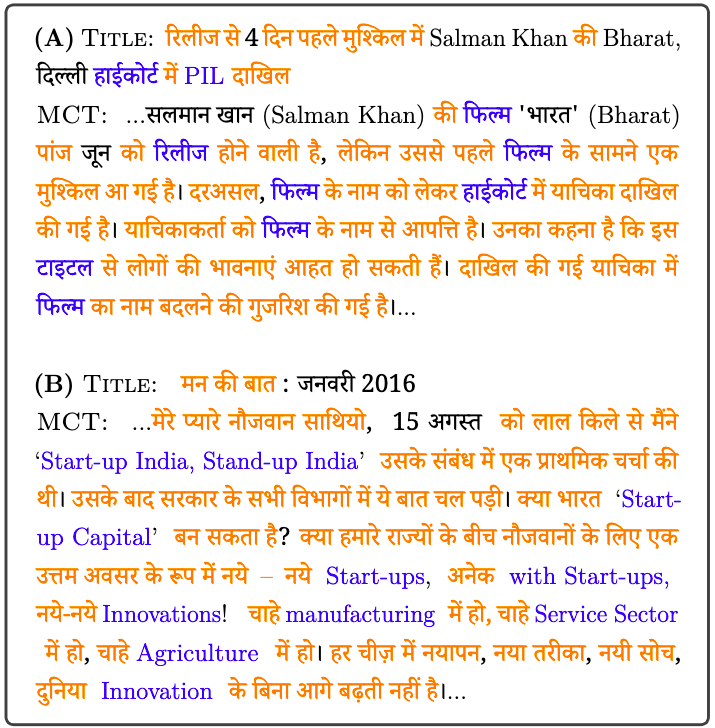}
\caption{Example MCT and the corresponding article's title form two multilingual data sources: (A) Dainik Jagran news article and (B) Man-ki-baat speech transcript. We color code the tokens as: \textcolor{blue}{English}, \textcolor{orange}{Hindi}, and language independent.}
\label{fig: example}
\end{figure}
\begin{table*}[!t]
\centering
\resizebox{\hsize}{!}{
\begin{tabular}{l|cccccc}
\textit{Dataset} & \textit{Task(s)} & \textit{Data Source(s)} & \textit{\# Instances} & \textit{Avg Tokens} & \textit{Avg Sentences} & \textit{Retrieval} \\ \hline

% \cite{banerjee2018dataset} & \begin{tabular}[c]{@{}c@{}}Conversation \\ System\end{tabular} & \begin{tabular}[c]{@{}c@{}}Manual translation of\\ conversation text from\\ \cite{henderson2014second}\end{tabular} & 7091 & 8 & N/A \\ \hline

\cite{srivastava2020phinc} & \begin{tabular}[c]{@{}c@{}}Machine\\ Translation\end{tabular} & \begin{tabular}[c]{@{}c@{}}Social media posts\\on Twitter \& Facebook\end{tabular} & 13738 &  13 & 1.04 & Automatic  \\ \hline

\cite{khanuja2020new} & \begin{tabular}[c]{@{}c@{}}Natural Language\\ Inference\end{tabular} & \begin{tabular}[c]{@{}c@{}}Hindi Bollywood \\movie transcripts \end{tabular} & 2240 & 87 & 7.15 & Automatic  \\ \hline

\cite{mehnaz2021gupshup} &  \begin{tabular}[c]{@{}c@{}}Dialogue\\ Summarization\end{tabular} & \begin{tabular}[c]{@{}c@{}}Manual translation of\\ dialogues and summaries\\ from \cite{gliwa2019samsum} \end{tabular}& 6830 & 31 & 7.85 & - \\ \hline

\cite{srivastava2021hinge}        &    \begin{tabular}[c]{@{}c@{}}Generation \&\\ Evaluation\end{tabular}     &   \begin{tabular}[c]{@{}c@{}} IIT-B En-Hi parallel corpus\\ \cite{kunchukuttan2018iit}       \end{tabular}  &  1974  &  20 & 1.05 & -  \\ \hline

\textsc{MUTANT} & Summarization & \begin{tabular}[c]{@{}c@{}}Speech transcripts, press \\releases, and news articles\end{tabular}  & 84937 & 159 &  10.23 & \begin{tabular}[c]{@{}c@{}}Manual + \\Automatic \end{tabular}  \\

\end{tabular}}
\caption{Comparison of the \textsc{MUTANT} dataset with the currently available datasets in the Hinglish language.}

% Avg Len: average number of tokens (after rounding-off to the next integer) in the dataset. 
% Code-Mixed FT: filtering technique to identify the code-mixed text from the initially collected dataset. \\
% $^*$In this paper, we leverage the \textit{LongCoMix} dataset for the summarization task but the dataset has use cases in several other downstream tasks (see Section \ref{sec:summ}).

\label{tab: comparison}
\end{table*}

We posit that due to several inherent challenges, the NLP community hold back on building multi-sentential datasets for the low-resource and code-mixed languages. One of the most significant bottlenecks in building such resources is the unavailability of MCT on traditional and widely popular data sources such as social media platforms where the short-length and noisy code-mixed text is available in abundance. It presents several challenges such as the difficulty in curating a large-scale multi-sentential dataset at ease. Another major challenge is the lack of metrics to measure the degree of code-mixing in the multi-sentential framework. The existing metrics such as code-mixing index \cite{das2014identifying} and multilingual-index \cite{barnett2000lides} already suffers from major limitations \cite{srivastava2021challenges} in the short-length text format. In such a scenario, it gets mystifying to build a retrieval pipeline to identify MCT and we need to depend heavily on the expertise of human annotators which is a time and cost-demanding exercise. In this work, we address both of these challenges. As a representative use case, we base our work on Hinglish, a popular code-mixed language in the Indian subcontinent. But the insights from our exploration could be extended to other code-mixed language pairs. 

To address the first challenge, we identify two non-traditional multilingual data sources\footnote{these data sources have not been actively employed in building datasets for the code-mixed languages} i.e., political speeches and press releases along with Hindi daily news articles (discussed in detail in Section \ref{sec: dataset}). Figure \ref{fig: example} shows example Hinglish MCTs from two multilingual data sources. 
To address the second challenge, we propose a token-level language-aware pipeline and extend a widely popular metric (i.e., code-mixing index) measuring the degree of code-mixing in a multi-sentential framework. We demonstrate the effectiveness of the proposed pipeline with a minimal task-specific annotation which significantly reduces the overall human effort (discussed in detail in Section \ref{sec: setup}). 

Eventually, we build a novel multi-sentential dataset for the Hinglish language with 85k MCTs identified from 67k articles. In Table \ref{tab: comparison}, we compare \textsc{MUTANT} with four other Hinglish datasets \cite{srivastava2020phinc,khanuja2020new,mehnaz2021gupshup,srivastava2021hinge} proposed for a variety of tasks such as machine translation, natural language inference, generation, and evaluation. The \textsc{MUTANT} dataset has a significantly higher average number of sentences along with longer MCT (high average number of tokens). Alongside, the dataset notably consists of a higher number of data instances which is a rarity for the code-mixed datasets \cite{srivastava2021challenges}.
% To the best of our knowledge, we are one of the first works to employ a semi-automatic retrieval pipeline to identify the code-mixed text from the multilingual document. 

% The average number of tokens in a paragraph in the \textit{MUTANT} dataset is significantly higher than all the other previously available datasets. Also, the number of data instances in the \textit{MUTANT} dataset is manifold as compared to other widely-used and available code-mixed datasets. 

% Our contribution can be summarized as:
% \begin{itemize}[noitemsep,nolistsep,leftmargin=*]
%     \item We present a large-scale code-mixed long sequence dataset, \textit{MUTANT}.
%     \item We propose an automatic pipeline to extract the long-sequence code-mixed text from the publicly available data sources.
%     \item We highlight several use cases of the proposed dataset in Section \ref{sec:summ}. We further demonstrate the inefficacy and opportunities with the code-mixed abstractive summarization task.
% \end{itemize}

\section{Multi-sentential Code-mixed Text Span (MCT)}
\label{sec: MCT}
Due to the absence of a formal definition of MCT in the literature, we propose and use the following definition of MCT throughout this work: 

\noindent \underline{MCT}: Consider a multilingual article $A$ = \{$s_1$, $s_2$, ..., $s_n$\} consisting of $n$ sentences denoted by $s_i$ where $i\in$[1, $n$]. A unique non-overlapping MCT $M_p$ in $A$ is a chunk of $m>1$ consecutive sentences i.e. $M_p$ = \{$s_k$, $s_{k+1}$, ..., $s_{k+m-1}$\}. $M_p$ should satisfy the following two properties:
\begin{enumerate}[noitemsep,nolistsep,leftmargin=*]
    \item $P1$: At least one $s_{k+j}$ in $M_p$ should be code-mixed.
    % \footnote{a piece of text is code-mixed, if it contains words/phrases from at least two different languages}. 
    Trivially, at most $m$-1 $s_{k+j}$ in $M_p$ could be monolingual. Here, $j\in$[0, $m$-1].
    \item $P2$: $s_k$ in $M_p$ is either the first sentence of the article or preceded by a line break. Likewise, $s_{k+m-1}$ is either the last sentence of the article or succeeded by a line break.
\end{enumerate}

It should be noted that an article $A$ can have multiple non-overlapping unique MCTs i.e. $A$ = \{$M_1$, $M_2$, ..., $M_q$\} where $q\geq$0.

\section{Multilingual and Multi-sentential Data Sources}
\label{sec: dataset}
% \textcolor{red}{@RG: for both the datasets (speech and news), can you show some analysis of different topics and named-entities? You will need to use multilingual topic modeling and NER models.}
Over the years, we observe several interesting and diverse code-mixed data sources such as Twitter, Facebook, movie transcripts, etc.
Social media sites have acted as the cornerstone of the code-mixed data collection pipelines due to the ease of availability of large-scale data. Nonetheless, they present several challenges such as noisy data, short text, abusive, and multimodal data. Given the requirements of \textit{MUTANT} (i.e. multi-sentential and high-quality data), we refrain from using social media sites in this work. Here, we focus on two major data sources:

\subsection{Political speeches and press releases}
% \item \textbf{Political speeches and press releases, \boldsymbol{$D_{speech}$}:} 
Here, we scrape data from five different web sources. Collectively, we denote this data source as $D_{speech}$.

\noindent\textbf{Aam Aadmi Party press releases (\textsc{AAP})}: We scrape the press releases from the official website of Aam Aadmi Party\footnote{\url{https://aamaadmiparty.org/media/press-releases}}. We have scraped 320 Hindi press releases from their website. The website contains all the press releases in the last five years starting from June 2017.

\noindent\textbf{Indian National Congress speeches (\textsc{INC)}}: The official website of the INC stores some of the speeches by major INC political leaders. We have extracted 112 of these speeches from their official website\footnote{\url{https://www.inc.in/media/speeches}}. The timeline for the scraped speeches is between August 2018 to March 2022. 

\noindent\textbf{{Man-ki-baat} \textsc{(MKB)}}: Man-ki-baat is a radio program hosted by the Indian prime minister Narendra Modi where he periodically addresses the people of the nation. The MKB website\footnote{\url{https://www.pmindia.gov.in/hi/mann-ki-baat/}} stores the official transcripts in Hindi and English languages. We have extracted the transcripts of 67 of these programs between December 2015 to December 2021. 

\noindent\textbf{{Press Information Bureau \textsc{(PIB)}}}: The Press Information Bureau houses the official press releases from all Indian government ministries including President's office, the Prime Minister's office, Election Commission, etc. We have extracted 30283 articles from the PIB website\footnote{\url{https://www.pib.gov.in}}. The timeline for these articles is from June 2017 to March 2022.

\noindent\textbf{PM speech \textsc{(PMS)}}: Majority of the Indian Prime Minister speeches (different from MKB speeches) are stored digitally on the PM India website\footnote{\url{https://www.pmindia.gov.in/hi/news-updates/}}. We have extracted 694 of these speeches that are recorded between November 2016 to October 2021.

\subsection{Hindi news articles}
Here, we scrape data from two major Hindi news daily websites. Collectively, we denote this data source as $D_{news}$.

\noindent\textbf{Dainik Bhaskar \textsc{(DB)}}: Dainik Bhaskar is one of the most popular Hindi newspapers in India. It is ranked 4th in the world by circulation according to World Press Trends 2016\footnote{\url{https://web.archive.org/web/20170706110804/http://www.wptdatabase.org/world-press-trends-2016-facts-and-figures}}. They have digitized the daily newspapers on their website\footnote{\url{https://www.bhaskar.com}}. Articles on DB website have been divided into many categories such as `\textit{Entertainment}' and `\textit{Sports}'. We have extracted 115324 articles uploaded on the website between February 2019 to May 2022. In Table \ref{tab: news}, we present the category-wise distribution of the articles scraped from the DB website.

\begin{table}[!tbh]
\centering
\small
\begin{tabular}{l|cc}
    \textit{Category}          & \textit{DB}     & \textit{DJ}     \\ \hline

Business      & 16012  & 4203   \\ 
Entertainment & 18498  & 52173  \\ 
Featured      & 5536   & 19373  \\ 
Lifestyle     & 12189  & -      \\ 
Miscellaneous & 20221  & -      \\ 
National      & 18615  & 160005 \\ 
Politics      & -      & 33604  \\ 
Sports        & 9950   & -      \\ 
World         & 14303  & 42478  \\ 
Total         & 115324 & 311836
\end{tabular}
\caption{Number of articles in various news categories in the DB and DJ datasets.}
\label{tab: news}
\end{table}

\noindent\textbf{Dainik Jagran \textsc{(DJ)}}: Dainik Jagran is another popular Indian Hindi newspaper. According to World Press Trends 2016, DJ is ranked 5th in the world by circulation. Similar to the DB website, they have also created a repository of articles on their official website\footnote{\url{https://www.jagran.com}}. Here, we extract 311836 of these articles from the website that were uploaded between April 2013 to May 2022. In Table \ref{tab: news}, we present the category-wise distribution of the articles scraped from the DJ website.

\section{Experimental Setup}
\label{sec: setup}
\underline{\emph{Problem definition}}: Given a multilingual article $A$ comprising of $q$ multi-sentential text spans (MST) i.e. $A$ = \{$M_1$, $M_2$, ..., $M_q$\}, we predict a binary outcome $L_{CM}$ for each MST $M_i$ i.e. $L(A)$ = \{$L_{CM}^{M_1}$, $L_{CM}^{M_2}$, ..., $L_{CM}^{M_q}$,\}. $L_{CM}^{M_i}$ = 1, if $M_i$ is code-mixed, otherwise 0. In a nutshell, a code-mixed MST $M_i$ is a MCT and it satisfies the properties $P1$ and $P2$ (ref. \cref{sec: MCT}).

\begin{figure*}[!tbh]
    \centering
    \includegraphics[width=1.0\linewidth]{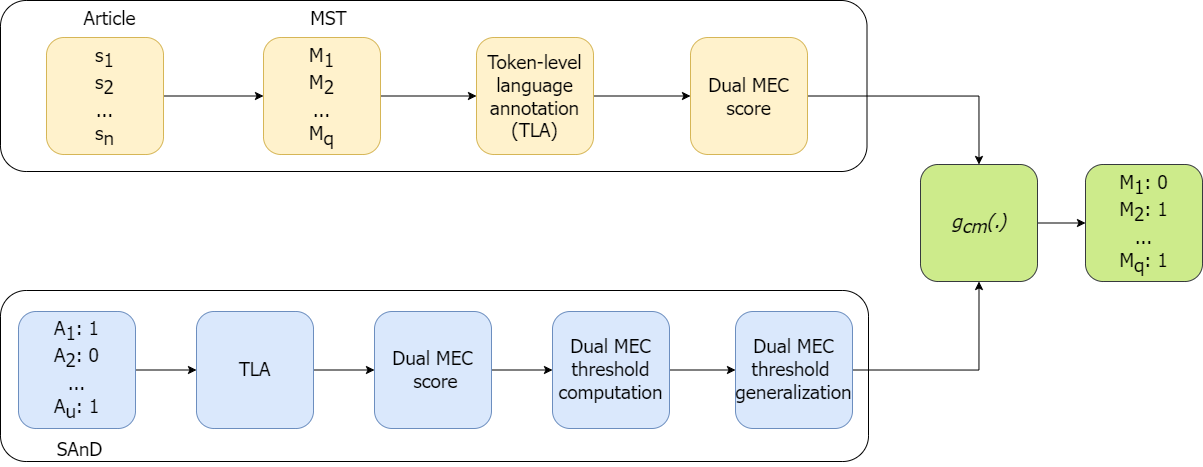}
    \caption{Architecture of MCT identification pipeline.}
    \label{fig: arch}
\end{figure*}

Figure \ref{fig: arch} shows the architecture of the MCT identification pipeline. Next, we discuss the various components of this pipeline in detail.

% \noindent\underline{\emph{Notations}}:
% MST $M_i$ consists of $pi$ words i.e. $M_i$ = [$w_{1i}$, $w_{2i}$, ..., $w_{pi}$], where $w_{ji}$ represents $j$th word in $M_i$ and $pi$ need not equal to $pj$, if $i$\neq$j$. Also, $M_i$ consists of $ki$ sentences i.e. $M_i$ = [$s_{1i}$, $s_{2i}$, ..., $s_{ki}$], where $s_{ji}$ represents $j$th sentence in $M_i$ and $ki$ need not equal to $kj$, if $i$\neq$j$.

\subsection{Token-level language annotation (TLA)}
We exploit the token-level language information to identify MCT given a multilingual article $A$. We annotate the words in $A$ using a code-mixed language identification tool. Specifically, we use L3Cube-HingLID \cite{nayak2022l3cube} for this task. A word $w_i$ $\in$ $A$ can take either of the three language tags from the set \{$English$, $Hindi$, $Other$\}.  Given that L3Cube-HingLID works only on the Roman script text, we use a Devanagari to Roman script transliteration tool\footnote{\url{https://github.com/ritwikmishra/devanagari-to-roman-script-transliteration}} for the tokens written in Devanagari script.
In Table \ref{tab: scrped_data}, we report the percentage of $Hindi$ and $English$ tokens. With an exception of the AAP dataset, $Hindi$ is the predominant language in all the data sources. 
\begin{table}[hbt!]
\resizebox{\hsize}{!}{
\begin{tabular}{l|ccccc}
\textit{}  & \textit{Articles} & \textit{AW} & \textit{AC} & \textit{\%H} & \textit{\%E}\\ \hline
AAP              & 320                                             & 1129                                             & 6033     &   53.97 & 45.09   \\ 
INC & 112                                             & 2312                                           & 10691    &    63.83 & 33.12                                          \\ 

MKB      & 67                                              & 4151                                             & 20706       &      77.17 & 22.41                 \\ 

PIB              & 30283                                           & 525                                              & 3015        &     80.96 & 17.59                                      \\ 

PMS        & 694                                             & 2591                                             & 13400       &   79.02 & 20.45                                        \\

DB   & 115324                                          & 382                                             & 1977                &   80.22 & 18.25                               \\ 
DJ    & 311836                                       & 391                                               & 2037              &       79.28 & 19.60                                \\ 
$D_{speech}$        &  31476                                            &   590                                           &  3339      & 79.97   &   18.65                                    \\  
$D_{news}$        & 427160                                              &     388                                         &   2020     &   80.18    & 18.51                                   \\  
\begin{tabular}[c]{@{}c@{}}$D_{speech}$ \\ + $D_{news}$\end{tabular}        &      458636                                        &      401                                        &  589       & 80.05  &  18.54                                       \\   
\end{tabular}}
\caption{Distribution of the scraped articles from various data sources. AW: average number of words. AC: average number of characters. \%E: percentage of English tokens. \%H: percentage of Hindi tokens.}
\label{tab: scrped_data}
\end{table}

\subsection{Code-Mixing Index (CMI)}
In the literature, we observe several metrics that has been proposed to measure the degree of code-mixing in text such as code-mixing index (CMI, \cite{das2014identifying}), multilingual-index (M-index, \cite{barnett2000lides}) and integration-index (I-index, \cite{guzman2017metrics}). Each of these metrics has its own merits and limitations \cite{srivastava2021challenges}. In this work, we use the most widely used CMI metric due to the ease of interpretation and the suitability for the task. CMI, by definition, measures the degree of code-mixing in a text as:
\begin{equation}
\label{eq:oldCMI}
    CMI= \begin{cases}
100 * [1- \frac{max(w_{i})}{n-u}] & n> u \\
0 & n=u
\end{cases}
\end{equation}

Here, $w_{i}$ is the number of words of the language $i$, max\{{$w_{i}$}\} represents the number of words of the most prominent language, $n$ is the total number of tokens, $u$ represents the number of language-independent tokens (such as named entities, abbreviations, mentions, and hashtags). The CMI score ranges from 0 to 100. A low CMI score suggests the prevalence of only one language in the text whereas a high CMI score indicates a high degree of code-mixing.   

\subsection{Small annotated dataset (SAnD)}
We create a small manually annotated dataset comprising all seven data sources. The objective of the annotation is to assign a binary label to each MST such that we can identify if the MST is code-mixed or not from the assigned label.

More formally, $SAnD$ = \{$A_1$: $l_1$, $A_2$: $l_2$, ..., $A_u$: $l_u$\}, represents $u$ manually annotated MST\footnote{For distinctive representation, we denote MST in $SAnD$ with $A$ instead of $M$.} where $l_i\in$\{0,1\} $\forall$ $i\in$[1,$u$]. Here, $l_i$=1, if $A_i$ is code-mixed, otherwise 0.

For this annotation task, we have selected a small number of articles (60 each from $D_{speech}$ and $D_{news}$) randomly from the scraped articles. We leave it to the judgment of the annotator to decide if a sentence (and subsequently the MST) is code-mixed or not. The annotator has expert-level proficiency in Hindi, English, and Hinglish languages. In Table \ref{tab: SAnD}, we show the distribution of the annotated articles for each data source. In total, we annotate 120 articles and 568 MST where we identify 121 MST (21.3\%) as code-mixed.

% \usepackage{multirow}
\begin{table}[]
\centering
\small
\begin{tabular}{l|cccc}
\multirow{2}{*}{\textit{}} & \multirow{2}{*}{\textit{Articles}} & \multicolumn{3}{c}{\textit{MST}}                                     \\ 
% \cline{3-5} 
                         &                           & \multicolumn{1}{c}{Total} & \multicolumn{1}{c}{Hing} & E/H \\ \hline
AAP                      & 5                         & \multicolumn{1}{c}{6}     & \multicolumn{1}{c}{2}    & 4   \\
INC                      & 3                         & \multicolumn{1}{c}{69}    & \multicolumn{1}{c}{5}    & 64  \\
MKB                      & 3                         & \multicolumn{1}{c}{66}    & \multicolumn{1}{c}{25}   & 41  \\
PIB                      & 47                        & \multicolumn{1}{c}{62}    & \multicolumn{1}{c}{27}   & 35  \\
PMS                      & 2                         & \multicolumn{1}{c}{36}    & \multicolumn{1}{c}{13}   & 23  \\

DB                       & 30                        & \multicolumn{1}{c}{207}   & \multicolumn{1}{c}{48}   & 159 \\
DJ                       & 30                        & \multicolumn{1}{c}{122}   & \multicolumn{1}{c}{28}   & 94  \\

$D_{speech}$                  & 60                        & \multicolumn{1}{c}{239}   & \multicolumn{1}{c}{72}   & 167 \\

$D_{news}$                    & 60                        & \multicolumn{1}{c}{329}   & \multicolumn{1}{c}{76}   & 253 \\ 
\begin{tabular}[c]{@{}c@{}}$D_{speech}$ \\ + $D_{news}$\end{tabular}        & 120                       & \multicolumn{1}{c}{568}   & \multicolumn{1}{c}{148}  & 420 \\ 
\end{tabular}
\caption{$SAnD$ dataset statistics. Hing: Hinglish, E/H: English/Hindi.}
\label{tab: SAnD}
\end{table}

\subsection{Estimating multilinguality}
\label{sec: formulaion}
Though CMI is widely used in numerous previous works, we couldn't find any discussion on the ideal CMI score thresholding criteria to identify a good code-mixed text. The problem becomes even more challenging when we use the CMI metric in a  multi-sentential framework along with constraints $P1$ and $P2$ (ref \cref{sec: MCT}). Various works \cite{khanuja2020new} have used empirically identified CMI thresholds to measure the degree of code-mixing in the text. But, we couldn't find any experimental justification for their findings. 

\noindent \textbf{Dual MEC score}: Here, we propose a novel adoption of the CMI metric in a constrained multi-sentential framework. For MST $M_p$ with $k$ sentences, we compute the scores for dual multilinguality estimation criteria (MEC) as:
\begin{enumerate}[noitemsep,nolistsep,leftmargin=*]
    \item Sentence-level CMI ($CMI$): We compute $CMI(s_i)$ for the sentence $s_{i}$$\in$ $M_p$ using the language-information of all the words in $s_{i}$ and the formulation given in \ref{eq:oldCMI}. 
    
    \item Multilinguality ratio ($MR$): We compute $C_{MR}$ for the MST $M_{p}$ as:
    
    \begin{equation}
    \label{eq: MR}
        MR (M_p) = \frac{N_{cm}}{k}
    \end{equation}
    Here, $N_{cm}$ and $k$ are the number of code-mixed and total sentences in $M_p$ respectively.
\end{enumerate}

\begin{figure}[!tbh]
    \centering
    \includegraphics[width=1\linewidth]{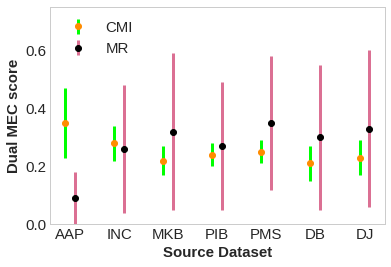}
    \caption{The mean and standard deviation of the dual MEC score for different data sources. The CMI score is scaled between 0 to 1.}
    \label{fig: mec_scores}
\end{figure}

\noindent Figure \ref{fig: mec_scores} shows the mean and standard deviation of dual MEC scores on seven different data sources.

\noindent \textbf{Formulation}: We identify if the sentence $s_i$ is code-mixed or monolingual using $CMI(s_i)$ score as:

\begin{equation}
\label{eq: fcm}
    f_{cm}(s_i) = \begin{cases}
1, & CMI(s_i) > \alpha \\
0, & otherwise
\end{cases}
\end{equation}

\noindent Here, $\alpha\in$ [0, 100] is the sentence-level CMI score threshold and $f_{cm}(.)$ estimates the code-mixing status ($1$ being code-mixed and $0$ being monolingual) of the sentence under consideration. Using \ref{eq: fcm}, we compute $N_{cm}$ as:

\begin{equation}
\label{eq: ncm}
    N_{cm} = \Sigma_{i=1}^{k} f_{cm}(s_i)
\end{equation}

% \begin{equation}
% \label{eq: nmono}
%     N_{mono} = k-\Sigma\limits_{i=1}^{k} f_{cm}(s_i)
% \end{equation}

\noindent Using \ref{eq: MR} and \ref{eq: ncm}, we compute $MR(M_p)$ as:
\begin{equation}
\label{eq: newMR}
     MR (M_p) = \frac{\Sigma_{i=1}^{k}f_{cm}(s_i)}{k}
\end{equation}

\noindent We formulate the following function to identify if MST $M_p$ with $k$ sentences is code-mixed:

\begin{equation}
    g_{cm}(M_p) = \begin{cases}
1, & MR(M_p) > \beta \\
0, & otherwise
\end{cases}
\end{equation}

\noindent Here, $\beta\in$[0, 1] is the multilinguality ratio threshold and $g_{cm}(.)$ estimates the code-mixing status ($1$ being code-mixed and $0$ being monolingual) of the MST under consideration. 

% We represent the CMI score of MST $M_p$ containing $k$ sentences using the dual CMI formulation as:
% \begin{equation}
% \label{eq:dualCMI}
%     CMI (M_p) = [\bigcup\limits_{i=1}^{k} C_{sent}^i; C_{MST}]
% \end{equation}

% \todo[inline]{@RG: Can you plot mean and standard-deviation of sentence-level CMI and MR using errorbar (https://jakevdp.github.io/PythonDataScienceHandbook/04.03-errorbars.html) for the original scraped dataset. X-axis will be all the 7 datasets (along with Dspeech and Dnews).We will have separate plots for CMI and MR.}

\subsection{Dual MEC threshold computation}
The dual MEC formulation helps us to identify the MCT in a constrained setting by jointly modeling the sentence-level and MST-level multilinguality information. As discussed in Section \ref{sec: formulaion}, the ideal thresholds $\alpha$ and $\beta$ are a conundrum that needs further exploration. Here, we propose to use the $SAnD$ dataset to identify the dual MEC thresholds ($\alpha$ and $\beta$). Algorithm \ref{algo:compute_thresh} shows the procedure to compute the thresholds. The algorithm takes $SAnD$ dataset $D$ with $u$ labeled MST. We represent the parameter search space for $\alpha$ and $\beta$ with $\alpha_{cand}$ and $\beta_{cand}$ respectively. $\alpha_{cand}$ ranges from $\alpha_{low}$ to $\alpha_{high}$ with a step-size of $\alpha_{step}$ whereas $\beta_{cand}$ ranges from $\beta_{low}$ to $\beta_{high}$ with a step-size of $\beta_{step}$. Based on our empirical observation, we set ($\alpha_{low}$, $\alpha_{high}$, $\alpha_{step}$) with (0, 50, 1) and ($\beta_{low}$, $\beta_{high}$, $\beta_{step}$) with (0, 0.5, 0.025).

\begin{figure}[!tbh]
    \centering
    \includegraphics[width=1\linewidth]{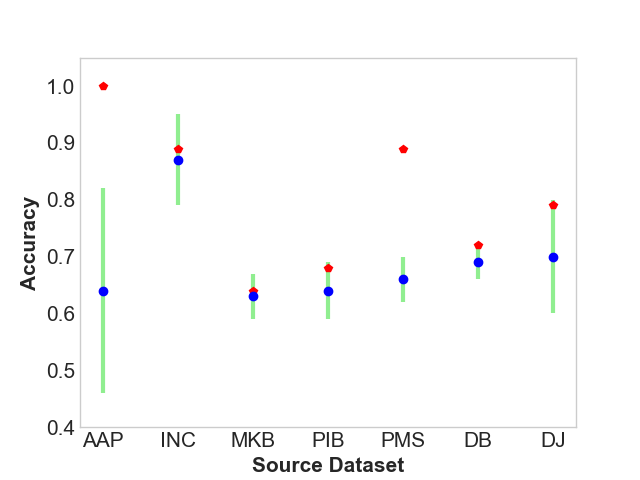}
    \caption{The mean and standard deviation of the accuracy on various dual MEC threshold combinations. The red dot corresponding to each data source indicates the accuracy against the best-identified thresholds.}
    \label{fig: accuracy}
\end{figure}

We perform the grid search on each threshold combination of ($\alpha_i$, $\beta_j$) to identify the best combination. For each threshold combination, we identify the accuracy of identifying the MCT in $D$ leveraging $f_{cm}(.)$ and $g_{cm}(.)$ formulations. We select the threshold combination with the highest accuracy as the final threshold ($\alpha$ and $\beta$). Table \ref{tab: thresholds} shows the best-identified thresholds on various data sources of the $SAnD$ dataset. Figure \ref{fig: accuracy} shows the mean and standard deviation of the accuracy on various dual MEC threshold combinations for different data sources.

\begin{algorithm}[]
\small
\caption{$compute_{\alpha,\beta}(D)$}\label{algo:compute_thresh}
\begin{algorithmic}[1]
\Require $D$ = \{$A_1$: $l_1$, $A_2$: $l_2$, ..., $A_u$: $l_u$\} where $A_i$ = \{$s_1$, $s_2$, ..., $s_k$\}
\Require $\alpha_{cand}$ = [$\alpha_{low}$, $\alpha_{low}$+$\alpha_{step}$, ..., $\alpha_{high}$]
\Require $\beta_{cand}$ = [$\beta_{low}$, $\beta_{low}$+$\beta_{step}$, ..., $\beta_{high}$]
\Require $Accuracy$ = \{\} 
% \Comment{key:value :: ($\alpha_i$, $\beta_j$):accuracy}

\For{$\alpha_i$ in $\alpha_{cand}$}
    \For{$\beta_j$ in $\beta_{cand}$}
        \State $hits$ = 0
        \For{$A_p$ $\in$ $D$}
            % \For{$s_q$ in $A_p$}
            %     \State Compute $f_{cm}(s_q)$
            % \EndFor
            \State $F_{cm}$ = $f_{cm}(s_q)$ $\forall$ $s_q$ $\in$ $A_p$ 
            \State Compute $g_{cm}(A_p)$ using $F_{cm}$ 
            % $f_{cm}(s_q) \forall s_q \in A_p$ 
            \If{$g_{cm}(A_p)$ == $l_p$}
                \State $hits$ = $hits$ + 1
            \EndIf
        \EndFor
    \State $Accuracy[(\alpha_i, \beta_j)] = 100*(hits/u)$
    \EndFor
\EndFor
\State $\alpha$ = $max_{value}(Accuracy).key()$[0]
\State $\beta$ = $max_{value}(Accuracy).key()$[1]

\State \Return $\alpha, \beta$
\end{algorithmic}
\end{algorithm}

\begin{table}[!tbh]
\centering
\small
\begin{tabular}{l|ccc}
\textit{}          & $\alpha$ & $\beta$ & \textit{Accuracy(\%)} \\
\hline
AAP             & 25                    & 0.35                 & 100          \\
INC             & 28                    & 0.30                 & 89           \\
MKB             & 22                    & 0.35                 & 64           \\
PIB             & 26                    & 0.15                 & 68           \\
PMS             & 21                    & 0.45                 & 89           \\

DB              & 18                    & 0.40                 & 72           \\
DJ              & 28                    & 0.40                 & 79           \\

$D_{speech}$         & 24                    & 0.35                 & 72           \\
$D_{news}$           & 29                    & 0.475                & 78           \\
\begin{tabular}[c]{@{}c@{}}$D_{speech}$ \\ + $D_{news}$\end{tabular}  & 29                    & 0.45                 & 75      \\    
\end{tabular}
\caption{Best identified thresholds ($\alpha$ and $\beta$) along with the accuracy of identifying MCT on various data sources in the $SAnD$ dataset.}
\label{tab: thresholds}
\end{table}

\subsection{Dual MEC threshold generalization}
As evident from Table \ref{tab: thresholds}, the thresholds $\alpha$ and $\beta$ vary across the data sources. So, it is important to identify which of these identified thresholds will result in a robust and stable performance across datasets. Here, we experiment with five dual MEC threshold generalisation techniques:

\begin{enumerate}[noitemsep,nolistsep,leftmargin=*]
    % \item \textbf{Random}: Here, we randomly sample five independent values for $\alpha$ $\in$ [0,100] and $\beta$ $\in$ [0,1] and set their average score as the dual MEC thresholds. 
    
    \item \textbf{Local Average (LA)}: For the data source $D_i$, we take the mean sentence-level CMI score and mean MR score as the dual MEC thresholds.
    
    \item \textbf{Global Average (GA)}: For the data source $D_i$, we take the mean sentence-level CMI score and mean MR score of the corresponding category data-source ($D_{speech}$ or $D_{news}$) as the dual MEC thresholds.
    
    \item \textbf{Average of LA and GA (ALG)}: For the data source $D_i$, we take the average of LA and GA identified thresholds as the dual MEC thresholds.
    
    \item \textbf{Single data source generalization (SDG)}: In this approach, we generalize the dual MEC thresholds identified locally on a single data source $D_i$ (using Algorithm \ref{algo:compute_thresh}) to identify MCT globally on other data sources.
    
    % For instance, to classify an unlabelled MST $M_p$ in the data source \textit{MKB}, we will use the thresholds $\alpha$ = 22 and $\beta$ = 0.35 (from Table \ref{tab: thresholds}) in the $f_{cm}(.)$ and $g_{cm}(.)$ formulations.
    
    \item \textbf{Multi data source generalization (MDG)}: In this approach, we use the dual MEC threshold information from multiple sources and use the majority voting to identify the best thresholds. For the data source $D_i$, we use the thresholds identified on three data sources (using Algorithm \ref{algo:compute_thresh}), namely $D_i$, $D_{speech}$ (if $D_i$ $\in$ $D_{speech}$, else  $D_{news}$), and $D_{speech} + D_{news}$. We then make an independent prediction on each of the three thresholds and take majority voting for the final classification of $M_p$. 
    % This formulation takes into account the  multilinguality information in the source dataset along with category-level and global-level datasets.
\end{enumerate}

\section{\textsc{MUTANT}: A \underline{Mu}lti-sen\underline{t}enti\underline{a}l Code-mixed Hi\underline{n}glish Da\underline{t}aset}
\label{sec: MUTANT}
We evaluate the performance of MCT identification pipeline and the five dual MEC threshold generalization techniques using the three subsets of the $SAnD$ dataset: $D_{speech}$, $D_{news}$, and $D_{speech} + D_{news}$. We report the following metric scores on each of the seven data sources:

\begin{enumerate}[noitemsep,nolistsep,leftmargin=*]
    \item Accuracy: We compute accuracy as the ratio of the total correct prediction of MCT and non-MCT to the total number of MST. We multiply this ratio by 100 and report the accuracy percentage. A high accuracy \% is preferred.
   
    \item False MCT Rate (FMR): We define FMR as the ratio of incorrectly identified MCT to the total number of actual monolingual MST. We report the FMR\% and a low FMR\% is preferred.
   
    \item Diversity@10 (D@10): We define D@10 as the percentage of articles in data source $D_i$ having more than $10\%$ correctly identified MCT. A high D@10 score is preferred.
\end{enumerate}

\begin{table}[]
\resizebox{\hsize}{!}{
\begin{tabular}{l|ccccc|ccccc|ccccc}
 & \multicolumn{5}{c|}{\textit{Accuracy}} & \multicolumn{5}{c|}{\textit{FMR}} & \multicolumn{5}{c}{\textit{D@10}} \\
 & L & G & A & S & M & L & G & A & S & M & L & G & A & S & M \\ \hline
AAP & 62 & 66 & 64 & 72 & 74 & 15 & 21 & 20 & 17 & 17 & 49 & 46 & 51 & 60 & 62 \\
INC & 63 & 66 & 64 & 73 & 74 & 17 & 21 & 20 & 16 & 12 & 49 & 46 & 51 & 59 & 59 \\
MKB & 61 & 66 & 62 & 69 & 72 & 28 & 21 & 26 & 22 & 18 & 51 & 46 & 48 & 68 & 70 \\
PIB & 62 & 66 & 64 & 67 & 72 & 24 & 21 & 24 & 30 & 17 & 53 & 46 & 55 & 73 & 74 \\
PMS & 67 & 66 & 64 & 71 & 74 & 17 & 21 & 23 & 20 & 16 & 51 & 46 & 53 & 67 & 69 \\
DB & 66 & 63 & 62 & 67 & 78 & 29 & 26 & 28 & 30 & 5 & 57 & 56 & 57 & 78 & 78 \\
DJ & 62 & 63 & 64 & 75 & 78 & 26 & 26 & 26 & 6 & 5 & 48 & 56 & 49  & 73 & 74
\end{tabular}}
\caption{Results on $D_{speech}$ dataset. L: LA, G: GA, A: ALG, S: SDG, M: MDG.}
\label{tab: speech_results}
\end{table}

We report the results in Tables \ref{tab: speech_results}, \ref{tab: news_results}, \ref{tab: news_speech_results}. The mean-based threshold generalization techniques (LA, GA, and ALG) consistently show poor performance on all the metrics. Given the nature of the problem, we prefer a low rate of misidentification of monolingual MST as the MCT and at the same time a high number of actual MCT should also be identified. MDG threshold generalization technique satisfies both conditions with low FMR and high accuracy on all the datasets. D@10 depicts if the threshold generalization technique is influenced by the presence of a few outliers in the dataset. SDG and MDG both show competitive results on the D@10 metric outperforming the mean-based threshold generalization techniques by a large margin. The constant poor performance of mean-based threshold generalization against SDG and MDG also shows the efficacy of the proposed threshold computation strategy (Algorithm \ref{algo:compute_thresh}).

Finally, to build the \textsc{MUTANT} dataset, we use the MCT identification pipeline with the MDG threshold generalization technique. Table \ref{tab: final_dataset} shows the statistics of the \textsc{MUTANT} dataset. To facilitate future work on this novel task of MCT identification, we will release the \textsc{MUTANT} dataset along with the initially scraped data from all the data sources and the annotated $SAnD$ dataset. The \textsc{MUTANT} dataset can be used for various tasks including but not limited to question-answering, text summarization and machine translation for Hinglish texts. This dataset could be used as a  pre-training dataset to train efficient NLU models for various tasks on Hinglish data.

\begin{table}[]
\resizebox{\hsize}{!}{
\begin{tabular}{l|ccccc|ccccc|ccccc}
 & \multicolumn{5}{c|}{\textit{Accuracy}} & \multicolumn{5}{c|}{\textit{FMR}} & \multicolumn{5}{c}{\textit{D@10}} \\
 & L & G & A & S & M & L & G & A & S & M & L & G & A & S & M \\ \hline
AAP & 72 & 70 & 71 & 72 & 73 & 17 & 15 & 17 & 14 & 14 & 60 & 58 & 62 & 70 & 72 \\
INC & 69 & 70 & 71 & 73 & 73 & 14 & 15 & 15 & 9 & 7 & 58 & 58 & 58 & 65 & 66 \\
MKB & 66 & 70 & 68 & 70 & 72 & 25 & 15 & 21 & 21 & 15 & 73 & 58 & 71 & 79 & 80 \\
PIB & 68 & 70 & 68 & 70 & 73 & 23 & 15 & 22 & 29 & 14 & 73 & 58 & 71 & 79 & 80 \\
PMS & 61 & 70 & 69 & 74 & 73 & 14 & 15 & 18 & 14 & 12 & 63  & 58 & 63 & 71 & 69 \\
DB & 66 &69 & 67 & 68 & 71  & 28 & 22 & 26 & 29 & 3 & 76  & 72 & 74 & 84 & 85 \\
DJ & 68 & 69 & 68 & 72 & 71 & 22 & 22 & 22 & 4 & 3 & 70  & 72 & 68  & 77 & 73
\end{tabular}}
\caption{Results on $D_{news}$ dataset. L: LA, G: GA, A: ALG, S: SDG, M: MDG.}
\label{tab: news_results}
\end{table}
\begin{table}[]
\resizebox{\hsize}{!}{
\begin{tabular}{l|ccccc|ccccc|ccccc}
 & \multicolumn{5}{c|}{\textit{Accuracy}} & \multicolumn{5}{c|}{\textit{FMR}} & \multicolumn{5}{c}{\textit{D@10}} \\
 & L & G & A & S & M & L & G & A & S & M & L & G & A & S & M \\ \hline
AAP & 69 & 70 & 69 & 73 & 74 & 12 & 15 & 15 & 13 & 13 & 55 & 60 & 57 & 65 & 66 \\
INC & 70 & 70 & 69 & 73 & 74  & 11 & 15 & 14 & 10 & 8 & 57 & 60 & 56 & 62 & 63 \\
MKB & 67 & 70 & 69 & 70 & 72  & 21 & 15 & 19 & 17 & 14 & 62 & 60 & 65 & 68 & 65 \\
PIB & 69 & 70 & 69 & 67 & 73  & 18 & 15 & 18 & 23 & 14 & 63 & 60 & 64 & 75 & 74 \\
PMS & 62 & 70 & 70 & 72 & 74 & 13 & 15 & 17 & 16 & 12 & 57 & 60 & 59 & 65 & 69 \\
DB & 67 & 68 & 67 & 67 & 75  & 23 & 19 & 22 & 24 & 4 & 64 & 62 & 62 & 76 & 75 \\
DJ & 68 & 68 & 69 & 74 & 75 & 19 & 19 & 19 & 5 & 4 & 57 & 62 & 62 & 71 & 74
\end{tabular}}
\caption{Results on $D_{speech}$+$D_{news}$ dataset. L: LA, G: GA, A: ALG, S: SDG, M: MDG.}
\label{tab: news_speech_results}
\end{table}

\section{Analysis and Discussion}
\label{sec: qual_eval}

\begin{table*}[]
\centering
\small
\begin{tabular}{l|c|c|c|ccc|ccc|ccc}
\multicolumn{1}{l|}{\multirow{2}{*}{\textit{\begin{tabular}[c]{@{}c@{}} \\\end{tabular}}}} & \multirow{2}{*}{\textit{A}} & \multirow{2}{*}{\textit{M}} &
\multirow{2}{*}{\textit{\begin{tabular}[c]{@{}c@{}}M/A\end{tabular}}} & 
\multicolumn{3}{c|}{\textit{Avg CMI}} &
\multicolumn{3}{c|}{\textit{Avg Words}}                                                            & \multicolumn{3}{c}{\textit{Avg Characters}}                                                       \\   &      &                                         &         &          A & M & H &
                                  A & M & H & A & M & H \\ \hline
AAP                               & 30                                    & 32                                      & 1.07                                              &     33.0    & 35.2   &   21.1                                    & \multicolumn{1}{c}{1347}             & \multicolumn{1}{c}{1263}                   &    16              & \multicolumn{1}{c}{6993}             & \multicolumn{1}{c}{6556}                   &    63             \\ 
INC                               & 85                                    & 306                                     & 3.6                                                &    28.1        &  27.5   &           -                  & \multicolumn{1}{c}{751}              & \multicolumn{1}{c}{208}                   &  -                & \multicolumn{1}{c}{3368}              & \multicolumn{1}{c}{935}                   &   -               \\ 
MKB                               & 58                                    & 243                                     & 4.19                                                 &   20.1          &   22.4  &                 -                 & \multicolumn{1}{c}{1034}              & \multicolumn{1}{c}{246}                   &   -               & \multicolumn{1}{c}{4843}             & \multicolumn{1}{c}{1156}                   &  -                \\ 
PIB                               & 8473                                  & 8786                                    & 1.04                                                 &    23.0       &    23.2 &                          21.0          & \multicolumn{1}{c}{572}              & \multicolumn{1}{c}{552}                   &          15        & \multicolumn{1}{c}{3139}             & \multicolumn{1}{c}{3028}                   &     87             \\ 
PMS                               & 597                                   & 3909                                    & 6.55                                                 &     25.8    & 24.7    &                         26.4             & \multicolumn{1}{c}{952}              & \multicolumn{1}{c}{145}                   &    13              & \multicolumn{1}{c}{4585}              & \multicolumn{1}{c}{700}                   &      79            \\ 
DB                                & 12851                                 & 15433                                   & 1.20                                                 &   21.0    &     21.2 &                      20.2                  & \multicolumn{1}{c}{107}               & \multicolumn{1}{c}{89}                   &      24            & \multicolumn{1}{c}{528}              & \multicolumn{1}{c}{440}                   &          123        \\ 
DJ                                & 44913                                 & 56228                                   & 1.25                                                 &  22.2    & 22.3    &                    21.6                     & \multicolumn{1}{c}{146}              & \multicolumn{1}{c}{117}                   &      16            & \multicolumn{1}{c}{734}              & \multicolumn{1}{c}{586}                   &      82            \\ 
$D_{speech}$                             &   9243                                    &              13276                           &   1.44                                              &    23.2    &  23.8   &      21.3                                      & \multicolumn{1}{c}{604}                 & \multicolumn{1}{c}{420}                   &     15             & \multicolumn{1}{c}{3258}                 & \multicolumn{1}{c}{2268}                   &         87         \\ 

$D_{news}$                             &      57764                                 &           71661                              &     1.24                    &    21.9 & 22.0                           &                  21.2                             & \multicolumn{1}{c}{137}                 & \multicolumn{1}{c}{111}                   &          18        & \multicolumn{1}{c}{688}                 & \multicolumn{1}{c}{555}                   &     91             \\ 

$D_{speech}$ + $D_{news}$                             &     67007                                  &        84937                                 &               1.27                 &    22.0  & 22.3                                                  &      21.2            & \multicolumn{1}{c}{201}                 & \multicolumn{1}{c}{159}                   &        17          & \multicolumn{1}{c}{1043}                 & \multicolumn{1}{c}{822}                   &           90       \\ 
\end{tabular}
\caption{\textsc{MUTANT} dataset statistics. A: Articles, M: MCT, and H: Headings. The INC and MKB datasets contain generic and very-low informative headlines and we do not include them in the final dataset.}
\label{tab: final_dataset}
\end{table*}
\begin{table}[]
\centering
\resizebox{\hsize}{!}{
\begin{tabular}{l|c|c|cc|c|ccc}
\multicolumn{1}{l|}{\multirow{2}{*}{\textit{\begin{tabular}[c]{@{}c@{}} \\ \end{tabular}}}}  & \multirow{2}{*}{\textit{A}} & \multirow{2}{*}{\textit{MST}} & \multicolumn{2}{c|}{\textit{CA}} & \multirow{2}{*}{\textit{CKS}} & \multirow{2}{*}{\textit{Acc}} & \multirow{2}{*}{\textit{FMR}} & \multirow{2}{*}{\textit{D@10}}\\ 
% \cline{4-6}
 &  &  & \multicolumn{1}{c}{Hing} & \multicolumn{1}{c|}{E/H} &\\ \hline
AAP & 5 & 5 & 2 & 3 &  1.0 & 100 & 0 & 100   \\ 
INC & 5 & 82 & 10 & 67 &  0.76 & 88 & 10 & 80 \\ 
MKB & 5 & 119 & 23 & 80 &  0.67 & 75 & 25 & 80\\ 
PIB & 5 & 5 & 2 & 3 &  1.0 & 80 & 0 & 50 \\ 
PMS & 5 & 141 & 13 & 110 & 0.52 & 84 & 12 & 100\\ 
DB & 5 & 49 & 3 & 43 & 0.63 & 78 & 20 & 50\\
DJ & 5 & 18 & 2 & 15 & 0.77 & 88 & 13 & 100\\ 
$D_{speech}$ & 25  & 352  & 50  &  263 & 0.65 & 82 & 14 & 71 \\ 
$D_{news}$ & 10 & 67 & 5  & 58 & 0.69 & 80 & 18 & 75 \\ 
\begin{tabular}[c]{@{}c@{}}$D_{speech}$ \\ + $D_{news}$\end{tabular} & 35 & 419 & 55 & 321 & 0.65 & 82 & 15 & 74 \\ 
\end{tabular}}
\caption{Qualitative evaluation of the \textsc{MUTANT} dataset. A: Articles, CA: complete agreement between the annotators, Hing: Hinglish MST. E/H: English/Hindi MST, CKS: Cohen's kappa score.}
\label{tab: evaluation}
\end{table}

In this section, we qualitatively evaluate the \textit{MUTANT} dataset by employing two human evaluators, different from the one used for the $SAnD$ to avoid any biases in the evaluation. Both evaluators are proficient in English, Hindi, and Hinglish languages. We randomly sample five articles from each of the seven source datasets and share the originally scraped articles containing both identified MCT and monolingual MST with both evaluators. During the evaluation, we do not disclose which of the MSTs is identified as MCT and share the following guidelines:

\begin{enumerate}[noitemsep,nolistsep,leftmargin=*]
     \item Any MST containing only Hindi words or only English words is monolingual.
    \item Any named entity, date, number, or word common in both English and Hindi languages should be considered a language-independent word.
\end{enumerate}

\begin{figure}[]
    \centering
    \includegraphics[width=1.0\linewidth]{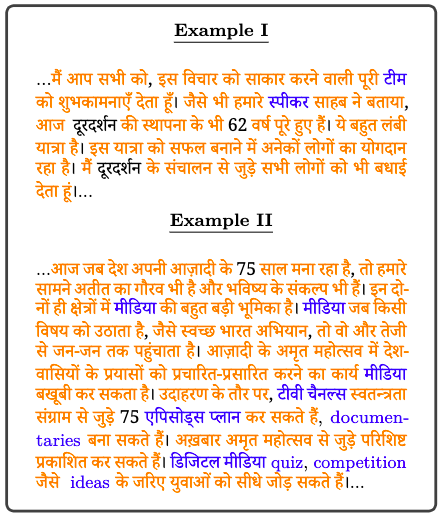}
    \caption{False positive MCT. We color code the tokens as: \textcolor{orange}{Hindi}, \textcolor{blue}{English}, and language independent.}
    \label{fig: fp_example}
\end{figure}

In Table \ref{tab: evaluation}, we report our findings from the qualitative evaluation study. Out of a total of 419 MST, we observe the complete agreement on 321 monolingual MST and 55 code-mixed MST resulting in $\approx$ 90\% complete agreement. A complete agreement means that both annotators agree that any particular MST is code-mixed or not. On MST with CA, we further compute the three metric scores using MDG. The results strengthen our earlier findings from Section \ref{sec: MUTANT}. In Figure \ref{fig: fp_example}, we report two example MCT incorrectly identified by our MCT identification pipeline. In the first example, both evaluators show complete agreement whereas in the second example there is a disagreement between the evaluators. We attribute this behavior to the poor state of the current code-mixed LID systems \cite{srivastava2021challenges} and since the CMI metric and our dual MEC formulation depend heavily on the code-mixed LID tools, the final results get affected. This limitation further provides an opportunity for future works to explore the problem from different perspectives such as a token-level language-independent MCT identification pipeline. It will also be interesting to see how this pipeline performs with other code-mixed languages, especially in a low-resource setting.

\section{Conclusion}
In this paper, we present a novel task of identifying MCT from multilingual documents. We propose an MCT identification pipeline by extending CMI to the multi-sentential framework and leveraging the pipeline we build a dataset for the Hinglish language. We highlight several challenges in building such resources and our insights will be useful to future works in code-mixed and low-resource languages.

\section{Limitations}
The limitations with the \textit{MUTANT} dataset include but are not limited to:
\begin{itemize}[noitemsep,nolistsep,leftmargin=*]
    \item Contrary to the previous works, all the data sources comprises the non social media sites. This could potentially limit the diversity in the code-mixed text as observed on social media platforms.
   
    \item In the current form, the dataset is limited to only one code-mixed language. We believe the proposed technique to extract MCT could be expanded to other code-mixed languages in the future.
   
    \item The data sources could potentially have their own biases (topical, style of writing, etc). We expect future works to be cautious while generalizing the results obtained on this dataset.
\end{itemize}

\bibliography{anthology,custom}
\bibliographystyle{acl_natbib}
\end{document}